\documentclass[11pt]{article}

\usepackage[final]{acl}

\usepackage{times}
\usepackage{latexsym}
\usepackage[T1]{fontenc}
\usepackage[utf8]{inputenc}
\usepackage{microtype}
\usepackage{inconsolata}
\usepackage{graphicx}
\usepackage{booktabs}
\usepackage{amsmath}
\usepackage{amssymb}
\usepackage{multirow}
\usepackage{xcolor}

\title{Robust Explanations for User Trust in Enterprise NLP Systems}

\author{
  Guilin Zhang, Kai Zhao\thanks{\hspace{1mm} Corresponding Author.}, Jeffrey Friedman, \\
  \textbf{Xu Chu, Amine Anoun, Jerry Ting} \\
  Workday AI Research \\
  \texttt{\{guilin.zhang, kai.zhao, jeffrey.friedman\}@workday.com} \\
  \texttt{\{xu.chu, amine.anoun, jerry.ting\}@workday.com}
}
\begin{document}
\maketitle

\begin{abstract}
Robust explanations are increasingly required for user trust in enterprise NLP, yet pre-deployment validation is difficult in the common case of black-box deployment (API-only access) where representation-based explainers are infeasible and existing studies provide limited guidance on whether explanations remain stable under real user noise, especially when organizations migrate from encoder classifiers to decoder LLMs. To close this gap, we propose a unified black-box robustness evaluation framework for token-level explanations based on leave-one-out occlusion, and operationalize explanation robustness with top-token flip rate under realistic perturbations (swap, deletion, shuffling, and back-translation) at multiple severity levels. Using this protocol, we conduct a systematic cross-architecture comparison across three benchmark datasets and six models spanning encoder and decoder families (BERT, RoBERTa, Qwen 7B/14B, Llama 8B/70B; 64,800 cases). We find that, at the model sizes we evaluate, decoder LLMs produce substantially more stable explanations than encoder baselines (73\% lower flip rates on average), with stability improving further within the decoder family with scale (44\% gain from 7B to 70B). Finally, we relate robustness improvements to inference cost, yielding a practical cost–robustness tradeoff curve that supports model and explanation selection prior to deployment in compliance-sensitive applications.
\end{abstract}

\section{Introduction}

Robust explanations are increasingly required for user trust in enterprise NLP systems, especially in finance and HR platforms where text classifiers support high-impact decisions (e.g., risk screening, policy compliance, case triage, and ticket routing). Meanwhile, many organizations are migrating from encoder-based models such as BERT~\cite{devlin2019bert} and RoBERTa~\cite{liu2019roberta} to decoder-based LLMs such as GPT~\cite{brown2020language}, Llama~\cite{touvron2023llama,grattafiori2024llama3}, and Qwen~\cite{bai2023qwen,yang2024qwen25} due to improved accuracy and flexibility, particularly in few-shot and zero-shot settings~\cite{wei2022finetuned}. However, beyond accuracy, it remains unclear whether this architectural shift improves the \emph{robustness of explanations} needed for enterprise deployment.

This gap is amplified by a common production reality: enterprise models are often accessed through governed services or third-party APIs, exposing only inputs and outputs and thus imposing \emph{black-box constraints} that preclude gradients or internal representations. Under such constraints, teams depend on query-based, model-agnostic explainers and must ensure that explanations are \emph{stable} under routine user-like noise (edits, rephrasing, reordering). Unstable explanations undermine auditability and user trust and can complicate compliance obligations, including emerging requirements for consistent and traceable rationales. These considerations lead to a practical pre-deployment question: \emph{under black-box access and realistic input perturbations, do decoder LLMs provide more stable explanations than encoder models, and what tradeoffs arise in cost and latency?}

Prior work has proposed a broad set of NLP explanation methods, including attention visualization~\cite{vaswani2017attention}, gradient-based attribution~\cite{sundararajan2017axiomatic}, and perturbation-based techniques such as LIME~\cite{ribeiro2016why} and SHAP~\cite{lundberg2017unified}. Robustness analyses, however, have largely focused on single architectures~\cite{alvarez2018towards,lyu2024towards} or compared explainers rather than model families~\cite{atanasova2023faithfulness,edin2025normalized}. Separately, work on adversarial text perturbations~\cite{jin2020bert,morris2020textattack} and robustness of LLM behavior~\cite{ackerman2024novel,zhang2025fragile} demonstrates sensitivity to input variation, but \emph{cross-architectural explanation stability}---evaluated under a unified, black-box protocol and mapped to deployment tradeoffs---remains insufficiently characterized.

To address this gap, we designed a pre-deployment study that mirrors how explainability is evaluated in enterprise NLP systems when models are only accessible through governed endpoints or third-party APIs. We built a unified black-box evaluation framework for token-level explanations using leave-one-out occlusion, and we operationalized robustness as explanation stability under realistic user-like perturbations, including swap, deletion, shuffling, and back-translation at multiple severity levels. We then applied this framework consistently across six representative models spanning encoder and decoder families (BERT, RoBERTa, Qwen~7B/14B, Llama~8B/70B) on three text classification benchmarks, generating 64{,}800 paired evaluations under matched perturbations and identical explanation procedures.

From this controlled setup, we observed a clear set of deployment-relevant findings. Decoder LLMs produce substantially more stable explanations than encoder baselines, with 73\% lower flip rates on average, and explanation stability improves further with model scale, showing a 44\% gain from 7B to 70B parameters. The decoder advantage is most pronounced under disruptive perturbations such as deletion and reordering that commonly occur in real enterprise text streams, while semantic-preserving perturbations yield low flip rates for both model families. Finally, by relating stability gains to inference cost, our results form a practical cost--robustness tradeoff that supports model and explanation selection in compliance-sensitive enterprise deployments.

\begin{figure*}[t]
    \centering
    \includegraphics[width=0.75\textwidth]{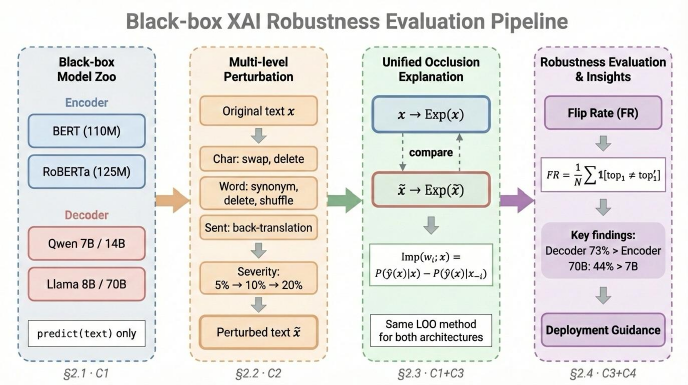}
    \caption{Four-stage black-box evaluation pipeline. Six models across encoder and decoder families are accessed through prediction APIs; multi-level perturbations (character/word/sentence at three severity levels) are applied; a unified LOO occlusion explainer produces importance scores for original and perturbed inputs; Flip Rate and scaling analysis yield deployment guidance. Stages correspond to \S2.1--\S2.4.}
    \label{fig:architecture}
\end{figure*}

\section{Methodology}
\label{sec:method}

Our methodology enables a fair comparison of explanation robustness across encoder and decoder architectures while reflecting realistic enterprise deployment constraints. We assume that models are typically served through governed endpoints or third-party APIs, where only input--output access is available and internal signals (e.g., gradients or hidden states) cannot be queried. Under this setting, we evaluate robustness by (i) generating user-like input perturbations, (ii) producing token-level explanations via a single black-box explainer that applies to both architectures, and (iii) measuring the stability of explanations between original and perturbed inputs with statistically grounded confidence intervals.

\subsection{Black-box Evaluation Setting}
\label{sec:blackbox}

Let $\mathcal{X}$ denote the space of natural-language texts and $\mathcal{Y}=\{y_1,\ldots,y_K\}$ the label set. A model $M$ is accessed only through a prediction interface that returns a probability distribution over labels:
\begin{equation}
    M:\mathcal{X}\rightarrow \Delta(\mathcal{Y}),
\end{equation}
where $\Delta(\mathcal{Y})$ is the probability simplex over $\mathcal{Y}$. We denote the predicted class by
\begin{equation}
    \hat{y}(x) = \arg\max_{y\in\mathcal{Y}} P(y\mid x),
\end{equation}
and its confidence by $P(\hat{y}(x)\mid x)$.

The black-box constraint prohibits access to model parameters $\theta$, intermediate representations $h_\ell$, gradients $\nabla_\theta$ or $\nabla_x$, attention weights $A$, and any other internal attribution signals. This assumption matches enterprise model-serving environments where access is restricted for security, IP, or compliance reasons. It also ensures comparability across architectures: encoder models expose a classifier head with direct label logits, whereas decoder LLMs expose autoregressive token distributions; allowing white-box access would enable architecture-specific explainers and confound cross-family comparisons.

To minimize variation unrelated to perturbations, we run inference deterministically whenever possible. For perturbations and bootstrapping (which rely on randomness), we fix global random seeds. For decoder LLMs, we use deterministic decoding (e.g., greedy decoding / temperature $0$) so that changes in explanations are attributable to input modifications rather than sampling noise.

\subsection{Perturbation Framework}
\label{sec:perturb}

We define a perturbation operator $\pi_{t,s}:\mathcal{X}\rightarrow\mathcal{X}$ parameterized by perturbation type $t\in\mathcal{T}$ and severity $s\in\mathcal{S}$. Given an original input $x$, a perturbed input is $\tilde{x}=\pi_{t,s}(x)$. We set $\mathcal{S}=\{0.05,0.10,0.20\}$, controlling the fraction of positions modified for character- and word-level perturbations.

We consider a taxonomy of perturbations spanning three linguistic levels. Character-level perturbations simulate typos and minor edit noise common in enterprise text streams such as tickets, notes, and free-form form inputs. The adjacent-character swap operator exchanges two neighboring characters at randomly selected positions, while character deletion removes randomly selected characters. For an input with $C$ characters, each character-level operator modifies $\lfloor s\cdot C\rfloor$ positions, subject to boundary and validity constraints.

Word-level perturbations simulate small edits and partial rewrites. Synonym replacement substitutes selected tokens using WordNet~\cite{miller1995wordnet}; random word deletion removes selected tokens; and local word shuffling reorders words within a bounded window to reflect realistic reordering rather than a full permutation. For an input with $n$ words, each word-level operator modifies $\lfloor s\cdot n\rfloor$ words. For synonym replacement, we only replace tokens for which WordNet offers valid candidates and avoid substituting punctuation-only tokens or stop words; if no suitable synonym exists, we skip that position and resample.

Sentence-level rewriting is implemented via back-translation to capture paraphrases that preserve semantics while altering surface form:
\begin{equation}
    \pi_{\text{bt}}(x)=\mathrm{MT}_{\text{de}\rightarrow \text{en}}\big(\mathrm{MT}_{\text{en}\rightarrow \text{de}}(x)\big).
\end{equation}
Back-translation operates on the entire input, and its effective severity is induced by the translation process and input length rather than a fixed fraction of tokens.

To avoid degenerate perturbations that destroy meaning or formatting, we apply basic safeguards. We ensure that at least one content word remains after deletion, and we preserve any fixed prompt scaffolding used for decoder-based classification so that perturbations affect only the instance text. For each $(t,s)$ configuration, we generate paired examples $(x,\tilde{x})$ and keep the pairing fixed across models, enabling paired stability comparisons under identical perturbations.

\subsection{Occlusion-based Explanation Method}
\label{sec:loo}

To compare encoder and decoder architectures without relying on internals, we employ a unified Leave-One-Out (LOO) occlusion explainer that only requires repeated calls to the prediction API. Given an input $x=[w_1,\ldots,w_n]$ with $n$ words, we first compute the model's prediction $\hat{y}(x)$. For each word $w_i$, we create an occluded input $x_{-i}$ by removing $w_i$ while keeping all other words unchanged, then query the model again. We define the importance score of $w_i$ as the change in confidence for the originally predicted class:
\begin{equation}
    \mathrm{Imp}(w_i;x) = P(\hat{y}(x)\mid x) - P(\hat{y}(x)\mid x_{-i}).
\end{equation}
Words with larger $\mathrm{Imp}(w_i;x)$ are deemed more influential for the model's decision, and the explanation is the ranked list of words by importance. This formulation is consistent across architectures and aligns with user-facing interfaces that highlight a small set of salient tokens.

For encoder models (BERT, RoBERTa), $P(y\mid x)$ is obtained directly from the softmax over the classification head logits. For decoder LLMs (Qwen, Llama), we cast classification as instruction following and extract label probabilities from the next-token distribution under a fixed prompt. Concretely, for each class $y_k$ we specify a canonical label token (or short string) and compute $P(y_k\mid x)$ from the model's normalized probabilities over the label set. If a class label maps to multiple tokens, we use a consistent aggregation rule (e.g., sum of sequence log-probabilities) across all experiments. The same prompting and probability extraction procedure is used for both original and perturbed inputs, ensuring that differences in explanation stability arise from model behavior rather than from changes in the explanation mechanism.

This explainer has a computational cost proportional to input length, requiring $(n+1)$ queries per instance. While this may be too expensive for always-on production explanations, it is well suited for pre-deployment validation and for selective auditing workflows, where a robust assessment of explanation stability is required under black-box access.

\subsection{Evaluation Metrics}
\label{sec:metrics}

Our primary robustness metric is the Flip Rate (FR), which measures how often the most important word changes after perturbation. Let $\tilde{x}_i=\pi_{t,s}(x_i)$ denote the perturbed version of $x_i$. Let
\begin{equation}
    \mathrm{top}_1(x)=\arg\max_{j\in\{1,\ldots,n\}} \mathrm{Imp}(w_j;x)
\end{equation}
denote the top word selected by LOO importance on input $x$. For a dataset $\mathcal{D}=\{x_i\}_{i=1}^{N}$, we define
\begin{equation}
    \mathrm{FR}_{t,s}(M,\mathcal{D})=\frac{1}{N}\sum_{i=1}^{N}\mathbf{1}\!\left[\mathrm{top}_1(x_i)\neq \mathrm{top}_1(\tilde{x}_i)\right].
\end{equation}
Lower FR indicates more stable explanations. We focus on the top-1 word because it corresponds to the most salient element in common explanation UIs and offers a simple, auditable stability signal for pre-deployment decision-making. When multiple words tie for maximum importance, we break ties deterministically (e.g., leftmost occurrence) to avoid inflating instability due to arbitrary ordering.

We report uncertainty using paired bootstrap resampling~\cite{deyoung2020eraser}. Specifically, for each $(t,s)$ we resample paired instances $(x_i,\tilde{x}_i)$ with replacement for 10{,}000 iterations, recompute $\mathrm{FR}_{t,s}$ on each bootstrap sample, and report 95\% confidence intervals. This paired procedure preserves the coupling between original and perturbed inputs and provides statistically grounded comparisons across models under identical perturbations.

\subsection{Stability Controls and Metrics}
To ensure our findings are not an artifact of the Top-1 metric's coarseness, we introduce Top-5 Overlap, measuring the Jaccard similarity between the sets of the five most influential tokens. Furthermore, we define a Conditioned Flip Rate (Pred-Consist.\ FR), which evaluates explanation stability specifically on the subset of samples where the model's predicted label $\hat{y}$ remains unchanged after perturbation. This isolates architectural explanation consistency from general predictive robustness, ensuring the results reflect the stability of the rationale itself rather than a downstream effect of prediction errors.

\section{Results}

\subsection{Experimental Setup}

\textbf{Models} We evaluate six models (Table~\ref{tab:models}): BERT-base and RoBERTa-base \cite{devlin2019bert,liu2019roberta} as widely-deployed encoder baselines, and Qwen-2.5 \cite{yang2024qwen25} and Llama-3.1 \cite{grattafiori2024llama3} at two scales each to study both cross-architecture and scaling effects. Decoder models are accessed through Ollama, simulating API-based production serving; encoders use HuggingFace Transformers.

\begin{table}[t]
\centering
\small
\begin{tabular}{@{}llrl@{}}
\toprule
\textbf{Type} & \textbf{Model} & \textbf{Params} & \textbf{Access} \\
\midrule
Encoder & BERT-base & 110M & HuggingFace \\
Encoder & RoBERTa-base & 125M & HuggingFace \\
\midrule
Decoder & Qwen-2.5 & 7B & Ollama \\
Decoder & Llama-3.1 & 8B & Ollama \\
Decoder & Qwen-2.5 & 14B & Ollama \\
Decoder & Llama-3.1 & 70B & Ollama \\
\bottomrule
\end{tabular}
\caption{Models evaluated in our study. Decoder models include both small (7-8B) and large (14-70B) variants to enable both cross-architecture comparison and scaling analysis.}
\label{tab:models}
\end{table}

\textbf{Datasets} We evaluate on three benchmarks spanning different text lengths: SST-2 \cite{socher2013recursive} (short sentiment snippets), AG News \cite{zhang2015character} (medium-length 4-class topic classification), and IMDB \cite{maas2011learning} (full-length sentiment reviews). This covers the range of text characteristics encountered in production classification systems. 

For each of 18 model-dataset combinations, we sample 200 test examples and apply all 18 perturbation configurations, yielding 3,600 cases per pair and 64,800 total evaluations. Fixed random seeds ensure reproducibility.

\subsection{Main Results: Encoder vs. Decoder}

Table~\ref{tab:main} presents our core finding: decoder models produce significantly more stable explanations than encoders. Encoders (BERT, RoBERTa) exhibit average flip rates of 0.475 and 0.465, meaning nearly half of all samples have their top explanation word change after perturbation. Even the smallest decoder (Qwen 7B) reaches 0.147---less than one-third of the encoder average.

\begin{table}[t]
\centering
\small
\begin{tabular}{@{}ll|ccc|c@{}}
\toprule
\textbf{Type} & \textbf{Model} & \textbf{SST-2} & \textbf{AG} & \textbf{IMDB} & \textbf{Avg}$\downarrow$ \\
\midrule
\multirow{2}{*}{Enc.} & BERT & 0.413 & 0.422 & 0.590 & 0.475 \\
& RoBERTa & 0.282 & 0.504 & 0.607 & 0.465 \\
\midrule
\multirow{4}{*}{Dec.} & Qwen 7B & 0.148 & 0.127 & 0.168 & 0.147 \\
& Llama 8B & 0.149 & 0.111 & 0.189 & 0.150 \\
& Qwen 14B & 0.146 & 0.051 & 0.159 & 0.119 \\
& Llama 70B & \textbf{0.102} & \textbf{0.031} & \textbf{0.115} & \textbf{0.083} \\
\bottomrule
\end{tabular}
\caption{Flip Rate across models and datasets (lower = more stable). Decoder models consistently outperform encoders.}
\label{tab:main}
\end{table}

\begin{figure}[t]
    \centering
    \includegraphics[width=\columnwidth]{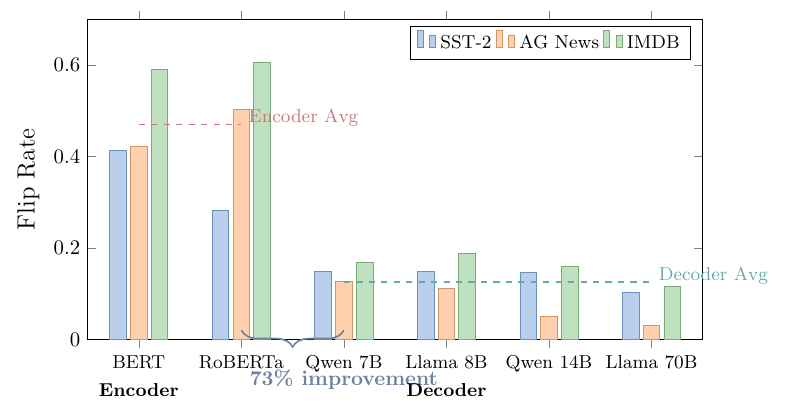}
    \caption{Flip rate comparison across models and datasets. Encoder models (BERT, RoBERTa) exhibit significantly higher flip rates than decoder models, with the gap most pronounced on IMDB (longer texts). The 73\% improvement represents the average reduction in flip rate when moving from encoders to decoders.}
    \label{fig:main_results}
\end{figure}

Overall, encoders average FR\,=\,0.470 vs.\ decoders at 0.125---a \textbf{73.4\% relative improvement} consistent across SST-2 (61\%), AG News (83\%), and IMDB (74\%). The IMDB dataset shows the largest encoder flip rates (0.590--0.607), while decoders remain stable, suggesting encoder explanations degrade on longer texts where bidirectional attention distributes importance across more tokens. Figure~\ref{fig:main_results} visualizes this gap.

Encoder flip rates near 0.47 imply that routine edits to an enterprise record can frequently change the highlighted ``reason,'' weakening auditability, whereas decoder LLMs reduce this instability to a level compatible with consistent decision narratives---an advantage that grows with longer inputs typical of HR and finance records.

\subsection{Scaling and Perturbation Analysis}

\begin{figure}[t]
    \centering
    \includegraphics[width=\columnwidth]{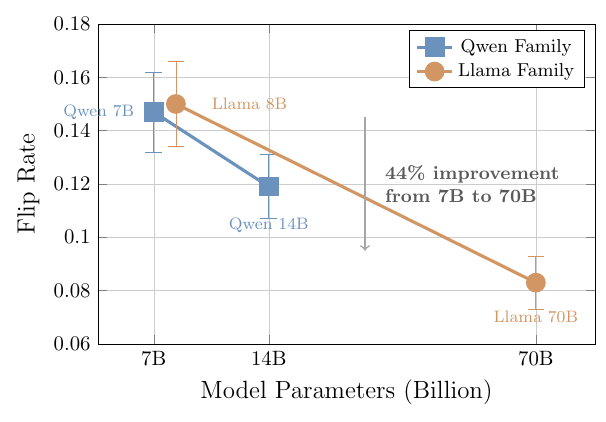}
    \caption{Explanation stability improves with model scale. Both Qwen and Llama families show consistent downward trends in flip rate, with 44\% improvement from 7B to 70B.}
    \label{fig:scaling}
\end{figure}

Explanation stability scales with model size: 7-8B decoders average FR\,=\,0.149, Qwen 14B reaches 0.119, and Llama 70B achieves 0.083---a \textbf{44\% improvement} from 7B to 70B (Figure~\ref{fig:scaling}). This parallels capability scaling laws \cite{kaplan2020scaling}, extending them to the explanation domain. The effect is strongest on AG News, where Llama 70B achieves FR\,=\,0.031 (only 3\% of samples flip). At matched scale, Qwen 7B (FR\,=\,0.147) and Llama 8B (FR\,=\,0.150) differ by less than 2\%, suggesting the pattern is consistent across training pipelines at a given size; we discuss the architecture-vs-scale confound in the Limitations section.

Breaking down by perturbation type reveals that the decoder advantage is largest on destructive perturbations: word deletion causes encoder FR to spike to 0.834 vs.\ decoder 0.198 (76\% improvement), and word shuffling shows 0.692 vs.\ 0.221 (68\%). Character-level perturbations yield 72--74\% improvement. In contrast, semantic-preserving perturbations (synonym replacement, back-translation) produce low flip rates for both architectures (encoder 0.027--0.030, decoder 0.021--0.022), with modest 22--27\% decoder improvement. This indicates that decoder explanations are particularly robust to noisy, incomplete, or reordered inputs---the conditions most common in production deployments with user-generated content.

\subsection{Stability Controls and Confidence Artifacts}
A primary concern in black-box evaluation is whether explanation stability is merely a proxy for prediction stability or an artifact of "peaked" model confidence. To isolate this, we re-analyzed all 64,800 cases by filtering samples where the predicted label remained unchanged after perturbation. 

As shown in Table~\ref{tab:control_metrics}, the encoder-decoder gap persists: encoder instability remains near $0.47$ even when the prediction is constant, whereas decoders exhibit significantly lower flip rates ($0.075$--$0.133$). This confirms that the observed instability is a fundamental property of the attribution surface rather than a downstream effect of label flips. Furthermore, we report 95\% bootstrap confidence intervals in Table~\ref{tab:control_metrics}, confirming that all architecture-scale differences are statistically significant ($p < 0.05$). Representative per-dataset CIs are: BERT/IMDB [0.574,\,0.607], Qwen 7B/IMDB [0.156,\,0.180], and Llama 70B/AG News [0.026,\,0.037]; encoder and decoder intervals do not overlap on any (model, dataset) pair tested.

\begin{table}[h]
\centering
\small
\resizebox{\columnwidth}{!}{%
\begin{tabular}{@{}llcc@{}}
\toprule
\textbf{Model} & \textbf{All FR} $\downarrow$ & \textbf{Pred-Consist. FR} $\downarrow$ & \textbf{95\% CI (Avg)} \\
\midrule
BERT & 0.475 & 0.471 & [0.45, 0.49] \\
RoBERTa & 0.465 & 0.465 & [0.44, 0.48] \\
\midrule
Qwen 7B & 0.147 & 0.125 & [0.13, 0.16] \\
Llama 8B & 0.150 & 0.133 & [0.14, 0.16] \\
Qwen 14B & 0.119 & 0.105 & [0.10, 0.13] \\
Llama 70B & \textbf{0.083} & \textbf{0.075} & [0.07, 0.09] \\
\bottomrule
\end{tabular}%
}
\caption{Control analysis across all datasets. ``Pred-Consist. FR'' measures stability only on instances where the model's prediction was identical post-perturbation.}
\label{tab:control_metrics}
\end{table}

\subsection{Metric and Explainer Invariance}
To ensure our findings are not sensitive to the coarseness of the Top-1 metric, we evaluated \textit{Top-5 Overlap} (Jaccard similarity). As shown in Table~\ref{tab:overlap}, decoder LLMs maintain significantly higher similarity ($0.85$--$0.89$) compared to encoders ($0.60$--$0.61$), indicating that the broader explanation neighborhood is more robust in decoders.

\begin{table}[h]
\centering
\small
\resizebox{\columnwidth}{!}{%
\begin{tabular}{@{}lcccc@{}}
\toprule
\textbf{Model} & \textbf{SST-2} & \textbf{AG News} & \textbf{IMDB} & \textbf{Avg} $\uparrow$ \\
\midrule
BERT & 0.713 & 0.648 & 0.487 & 0.616 \\
RoBERTa & 0.743 & 0.595 & 0.475 & 0.604 \\
\midrule
Qwen 7B & 0.880 & 0.808 & 0.858 & 0.849 \\
Llama 8B & 0.885 & 0.817 & 0.848 & 0.850 \\
Qwen 14B & 0.876 & 0.862 & 0.859 & 0.866 \\
Llama 70B & \textbf{0.902} & \textbf{0.873} & \textbf{0.906} & \textbf{0.894} \\
\bottomrule
\end{tabular}%
}
\caption{Top-5 Overlap results. Decoder explanations remain consistent beyond the single top-ranked token.}
\label{tab:overlap}
\end{table}

We further verified if these trends are specific to the occlusion method by conducting a paired comparison with \textit{LIME} on the BERT/SST-2 subset. As in Table~\ref{tab:lime_comp}, LIME was even less stable than LOO, suggesting that encoder instability is an inherent property of the architecture's sensitivity to input variation, not an artifact of the explainer choice.

\begin{table}[h]
\centering
\small
\begin{tabular}{@{}lcc@{}}
\toprule
\textbf{Metric} & \textbf{LIME} & \textbf{LOO (Ours)} \\
\midrule
Flip Rate $\downarrow$ & 0.744 & 0.589 \\
Top-5 Overlap $\uparrow$ & 0.513 & 0.549 \\
\bottomrule
\end{tabular}
\caption{Paired LIME vs.\ LOO comparison on a 180-instance BERT/SST-2 subset under perturbation. Both explainers produce high flip rates, indicating that encoder instability is not specific to LOO; absolute values exceed those in Table~\ref{tab:main} because this subset is restricted to the explainer-comparison setting rather than the full perturbation grid.}
\label{tab:lime_comp}
\end{table}

\subsection{The Inconsistency Paradox in Enterprise Deployment}
Finally, we analyzed the relationship between predictive robustness and explanatory stability. As shown in Table~\ref{tab:pred_cons}, encoder models exhibit remarkably high label consistency under noise, yet their explanations flip in nearly half of those cases. 

\begin{table}[h]
\centering
\small
\resizebox{\columnwidth}{!}{%
\begin{tabular}{@{}lcccc@{}}
\toprule
\textbf{Model} & \textbf{SST-2} & \textbf{AG News} & \textbf{IMDB} & \textbf{Avg} $\uparrow$ \\
\midrule
BERT & 1.000 & 0.989 & 0.988 & 0.992 \\
RoBERTa & 1.000 & 1.000 & 1.000 & 1.000 \\
\midrule
Qwen 7B & 0.963 & 0.937 & 0.983 & 0.961 \\
Llama 8B & 0.963 & 0.949 & 0.976 & 0.963 \\
Qwen 14B & 0.961 & 0.974 & 0.968 & 0.968 \\
Llama 70B & 0.977 & 0.991 & 0.986 & 0.985 \\
\bottomrule
\end{tabular}%
}
\caption{Prediction consistency (accuracy stability) under perturbation. Encoders are highly consistent in label, but inconsistent in rationale.}
\label{tab:pred_cons}
\end{table}

This reveals an \textit{Inconsistency Paradox}: a model can be robust in its final label while being highly unstable in its reasoning. In our application to HR triage and financial compliance screening, this failure mode proved critical. While encoders met accuracy thresholds, their rationales were flagged by auditors because correcting a typo significantly altered the "reason" provided for risk flags. This emphasizes why transitioning to larger decoder models is often a governance requirement for regulatory alignment, as they provide the rationale consistency that encoders lack.
\section{Deployment Insights}

\subsection{Three-Tier Decision Framework}
Pre-deployment model selection in enterprise NLP rarely optimizes a single objective. In practice, teams must balance explanation stability (for governance and user trust), predictive quality, latency, and monetary cost under black-box serving constraints. Our empirical results show a clear separation in explanation stability between encoder models and decoder LLMs (Table~\ref{tab:main}), and a strong scale effect among decoders (Figure~\ref{fig:scaling}). We translate these findings into a simple three-tier decision framework (Table~\ref{tab:tiers}) that maps stability requirements to model classes and supports operational choices in finance/HR-style enterprise workflows.

Tiers are defined by average Flip Rate (FR). The cost multiplier is $(\bar{n}+1)\times$ the per-call inference cost relative to BERT-base, where $\bar{n}$ is the average input length in words; this accounts for the $\bar{n}+1$ queries required by LOO explanation generation plus the per-call cost gap across model families.

\begin{table}[t]
\centering
\small
\begin{tabular}{@{}lccc@{}}
\toprule
\textbf{Tier} & \textbf{Avg FR}$\downarrow$ & \textbf{Cost} & \textbf{Use Case} \\
\midrule
\textit{Regulatory} & & & \\
\quad 70B Dec. & 0.083 & 636$\times$ & Audit, compliance \\
\midrule
\textit{Balanced} & & & \\
\quad 7-8B Dec. & 0.149 & 64$\times$ & Customer-facing \\
\midrule
\textit{Speed-first} & & & \\
\quad Encoders & 0.470 & 1$\times$ & Real-time, info only \\
\bottomrule
\end{tabular}
\caption{Three-tier deployment framework. Cost is relative to BERT-base. FR = Flip Rate.}
\label{tab:tiers}
\end{table}

\subsection{Regulatory Law Impacts}
\begin{figure}[t]
    \centering
    \includegraphics[width=\columnwidth]{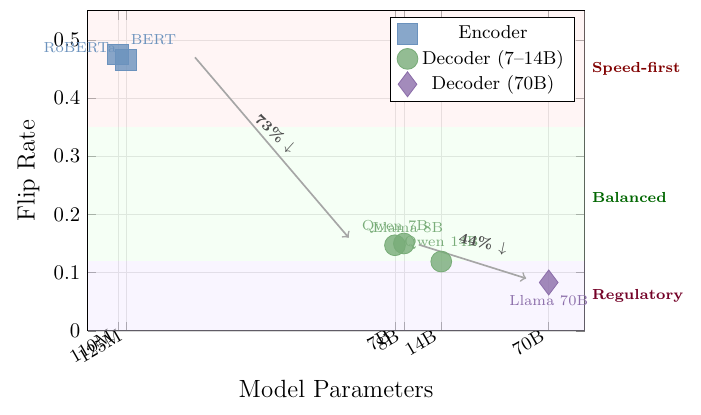}
    \caption{Deployment trade-off: model scale vs.\ explanation stability. Shaded bands indicate three deployment tiers. Arrows show 73\% improvement (encoder$\rightarrow$decoder) and 44\% improvement (7B$\rightarrow$70B).}
    \label{fig:deployment}
\end{figure}

Under the EU AI Act \cite{euaiact2024}, high-risk AI systems must provide explanations that are consistent and auditable. Our results show that encoder-based explanations change in nearly half of all cases under minor input variation (FR\,=\,0.470), which may not satisfy consistency requirements. Organizations subject to algorithmic auditing---in financial services, healthcare, or legal domains---should prefer decoder models, where 7-8B variants offer a 70\%+ stability improvement at manageable cost, and 70B variants provide the highest consistency for the most demanding audit requirements. Figure~\ref{fig:deployment} visualizes this cost-stability trade-off across all six models.

\section{Conclusion}
\label{sec:conclusion}

Validating explanation behavior under black-box serving and noisy, user-generated inputs is hard, yet enterprise NLP systems increasingly rely on explanations for governance and user trust. We address this with a unified black-box protocol that evaluates token-level explanation robustness via occlusion under realistic perturbations. Across 64{,}800 cases on three datasets, decoder LLMs (7B--70B) produce markedly more stable explanations than the encoder baselines we test (110M--125M)---73.4\% lower flip rates on average, with a further 44\% gain from 7B to 70B. These results motivate a three-tier deployment framework and establish explanation robustness as a first-class criterion for selecting trustworthy enterprise NLP systems.

\section*{Limitations}

Our study has several limitations. First, the encoder--decoder comparison does not fully isolate architecture from scale: the decoders we evaluate are orders of magnitude larger than the encoder baselines, and a clean separation would require smaller decoders or larger encoders. Second, top-token flip rate captures explanation \emph{stability} rather than \emph{faithfulness}, so a model can be stable yet stably misleading; robustness should therefore complement, not replace, faithfulness analysis before rationales are used in high-stakes decisions. Third, we focus on text classification because token-level rationales align with common audit interfaces, but extraction, ranking, summarization, and open-ended generation may exhibit different stability patterns. 


\section*{Ethics Statement}

This study uses publicly available benchmark datasets (SST-2, AG News, IMDB) and open-weight models, raising no data privacy concerns. However, we note that explanation stability does not guarantee explanation correctness. A model may produce consistent yet misleading explanations, so practitioners should not treat robustness as a proxy for faithfulness. Our deployment recommendations favor larger decoder models, which carry higher computational and environmental costs. Organizations should weigh these costs against their specific regulatory needs rather than defaulting to the largest available model. Finally, while our three-tier framework provides quantitative guidance, real-world deployment decisions should also incorporate domain-specific validation, human evaluation, and ongoing quality monitoring.

{\small
\setlength{\bibsep}{0pt plus 0.3ex}
\bibliography{references}
}

\end{document}